\documentclass[b4paper,10pt,twocolumn]{article}
\usepackage[utf8]{inputenc}
\usepackage[english]{babel}

\usepackage[margin={1.5cm,2.75cm}]{geometry}
\usepackage[font=small]{caption}
\usepackage{hyperref}
\usepackage{multirow}
\usepackage{url}
\usepackage{natbib}
\usepackage{amsmath,amsfonts,amssymb}
\usepackage{graphicx}
\usepackage{array}
\usepackage{booktabs}
\usepackage{multirow}
\usepackage{float}
\usepackage{siunitx}
\usepackage{subfig}

\let\oldtabular\tabular 
\renewcommand{\tabular}{\footnotesize\oldtabular}

\newcommand{\x}{\mathbf{x}}
\renewcommand{\v}{\mathbf{v}}

\newcommand{\dx}{\dot{\mathbf{x}}}

\newcommand{\y}{\mathbf{y}}

\newcommand{\R}{\mathbb{R}}
\newcommand{\0}{\mathbf{0}}
\newcommand{\bmu}{\boldsymbol\mu}

\newcommand{\m}{\mathbf{m}}

\newcommand{\bo}{\boldsymbol\omega}
\newcommand{\bs}{\boldsymbol\sigma}

\newcommand{\N}{\mathcal{N}}
\newcommand{\TN}{\mathcal{TN}}

\DeclareMathOperator{\diag}{diag}

\DeclareMathOperator*{\argmax}{arg\,max}

\newcommand{\dd}{  \:.\,.\, }

\begin{document}

\title{Bayesian Metabolic Flux Analysis reveals intracellular flux couplings}

\author{\large{Markus Heinonen$^{1,2}$, Maria Osmala$^{1}$, Henrik Mannerstr\"om$^{1}$, Janne Wallenius$^{3}$} \\ \large{Samuel Kaski$^{1,2}$, Juho Rousu$^{1,2}$ and Harri L\"ahdesm\"aki$^{1}$} \vspace{4mm} \\ 
$^{1}$Department of Computer Science, Aalto University, Espoo, 02150, Finland \\
$^{2}$Helsinki Institute for Information Technology, Finland\\
$^{3}$Institute for Molecular Medicine Finland, Helsinki, Finland}

\date{\vspace{-0ex}}

\twocolumn[
  \begin{@twocolumnfalse}
\maketitle

\begin{abstract}
\noindent\textbf{Motivation:} Metabolic flux balance analyses are a standard tool in analysing metabolic reaction rates compatible with measurements, steady-state and the metabolic reaction network stoichiometry. Flux analysis methods commonly place unrealistic assumptions on fluxes due to the convenience of formulating the problem as a linear programming model, and most methods ignore the notable uncertainty in flux estimates. \\
\textbf{Results:} We introduce a novel paradigm of Bayesian metabolic flux analysis that models the reactions of the whole genome-scale cellular system in probabilistic terms, and can infer the full flux vector distribution of genome-scale metabolic systems based on exchange and intracellular (e.g. 13C) flux measurements, steady-state assumptions, and target function assumptions. The Bayesian model couples all fluxes jointly together in a simple truncated multivariate posterior distribution, which reveals informative flux couplings. Our model is a plug-in replacement to conventional metabolic balance methods, such as flux balance analysis (FBA). Our experiments indicate that we can characterise the genome-scale flux covariances, reveal flux couplings, and determine more intracellular unobserved fluxes in \emph{C.~acetobutylicum} from 13C data than flux variability analysis. \\
\textbf{Availability:} The COBRA compatible software is available at \url{github.com/markusheinonen/bamfa} \\
\textbf{Contact:} \href{markus.o.heinonen@aalto.fi}{{markus.o.heinonen@aalto.fi}}  \\
\vspace{5ex}
\end{abstract}

  \end{@twocolumnfalse}
]

\section{Introduction}

Metabolic modelling considers networks of up to thousands of chemical reactions that transform metabolite molecules within cellular organisms \citep{palsson2015}. The key problem of metabolism is estimation of the reaction rates, or \emph{fluxes}, of the system of the highly interdependent intracellular fluxes from measurements of few exchange fluxes that transfer nutrients or products between the external medium and the cell.

The dominant approach to flux estimation is the celebrated Flux Balance Analysis (FBA) framework that finds reaction rates that maximise prespecified cellular growth function \citep{feist2010}, while assuming the cell is in a \emph{steady-state}, where concentrations of intracellular metabolites do not change \citep{almaas2004}. The FBA problem can be casted as a convenient and computationally efficient linear programming problem of solving a system of linear steady-state constraints while maximising a linear growth target \citep{orth2010}, and where flux measurements can be encoded as constraints to the fluxes \citep{carreira2014}. FBA is commonly used to characterise intra-cellular fluxes in various simulated target conditions \citep{mo2010}. In metabolic flux analysis (MFA) values of unknown fluxes are directly estimated based on measurements of some determined fluxes without explicit maximal growth assumption \citep{kim2008}. In both approaches a point estimate for up to thousands of highly interdependent fluxes are determined \citep{bordbar2014}.

The standard metabolic analyses suffer from several approximations that warrant careful methodological protocol to achieve biologically meaningful results. First, the exact steady-state constraint is an unrealistic assumption since metabolites can accumulate or deplete \citep{pakula2016}. Second, in FBA maximal growth is assumed, while it only holds at the highest growth phase in practise. Finally, due to a large number of metabolic reactions and limited number of experimental data, flux point estimates commonly used in the field completely ignore the notable uncertainty involved in FBA and MFA solutions. 
The flux variances are key in characterizing metabolic systems 
and uncertainties emerging from the use of insufficient and noisy data.

Numerous separate extensions to flux analysis have been introduced to alleviate these limitations. The robust FBA framework considers the effect of measurement uncertainties to the maximal growth \citep{zavlanos2011}. The steady-state assumption was recently relaxed by the RAMP model \citep{macgillivray2017}. In contrast to point flux estimates of FBA and MFA, the flux variability analysis (FVA) characterises the sensitivity of the target function to independent flux perturbations, resulting in upper and lower bounds around the FBA solution \citep{mahadevan2003,gudmundsson2010}. In principal flux mode analysis the eigenvectors of steady-state flux cone characterise the flux variability \citep{sahely2018}. Alternatively the solution space of the fluxes can be sampled \citep{schellenberger2010} by considering only optimal fluxes from border of the flux hypercone \citep{bordel2010} or by sampling also inoptimal fluxes from the inside the hypercone \citep{mo2010,saa2016}. The sampling methods use the 'Hit-and-Run' \citep{smith1984} or 'Artificially Centered Hit-and-Run' \citep{kaufman1998} algorithms to cope with the large flux space. A related approach uses possibility calculus \citep{dubois1996} to iteratively refine the estimate of possible and impossible flux states \citep{llaneras2009}.

There is a distinct lack of approaching metabolism with statistical modelling. The sole statistical approach for flux analysis remains the Metabolica framework that proposed modelling distributions of fluxes of skeletal muscle metabolism \citep{heino2007,heino2010}, but did not include modelling of maximal growths or genome-scale metabolic models. Bayesian methods have also been developed for 13C labeling data \citep{kadirkamanathan2006,theorell2017} by assuming fixed steady-state and without incorporating any target growth condition.

In this paper we tackle all three limitations simultaneously by introducing a novel paradigm of Bayesian metabolic flux analysis where the genome-scale, interdependent flux vector distributions are estimated. In Bayesian formalism prior distributions on flux random variables are determined, and subsequently updated by incorporating the measurement information, resulting in posterior flux distributions. We place priors on flux distributions, and estimate posterior distributions that characterise and quantify the probability of all flux states that are compatible with flux measurements, steady-state assumption and stoichiometry. Our model can reveal flux dependencies in explicit form, and characterise the full space of flux states in principled fashion. The Bayesian flux analysis can be used as a drop-in replacement to standard FBA, MFA, FVA and sampling methods. We provide public implementation of the Bayesian flux analysis using the standard COBRA framework \citep{cobra1,cobra2}.

\section{Methods}

The goal of this paper is a probabilistic formulation of static steady-state metabolic systems that can be applied to whole genome metabolic flux analysis (MFA), flux balance analysis (FBA) and flux variability analysis (FVA). We propose the Bayesian method as a direct replacement to these classic FBA, MFA and FVA tools. We start by assuming a metabolic system of $M$ metabolites and $N$ reactions has been characterised by a constant stoichiometric matrix $S \in \mathbb{Z}^{M \times N}$, where the rows denote metabolite participations in all reactions, while the columns denote reactants and products of metabolites by individual reactions. The flux vector $\v = (v_1, \ldots, v_N)^T \in \R^N$ denotes the reaction rates of the system. The \emph{steady-state} equation can be stated as 
$$S \v = \dx = \0,$$
which encodes that metabolite concentration changes $\dx \in \R^M$ are zero and hence the metabolite concentrations $\x \in \R^M$ do not change. Throughout the paper we assume a subset of fluxes have been observed or determined (for instance, some of the exchange fluxes), while the remaining fluxes are unknown. Our goal is to infer the distribution of all unknown fluxes given the observed fluxes, the steady-state constraints and the flux lower and upper bounds.

\subsection{Bayesian metabolic model}

We formulate a Bayesian flux model, which starts by assuming multivariate Gaussian priors for fluxes as
$$\v | \m_v, \bs_v \sim \N(\mathbf{m}_v, \Sigma_v),$$
with means $\m_v \in \R^N$ and diagonal covariances $\Sigma_v = \diag(\bs_v^2) = \diag (\sigma_{v_1}^2, \ldots, \sigma_{v_N}^2)$. The prior means are set to zero, or to the closest value to zero considering the flux upper and lower bounds. The variances $\sigma_{v_i}$ are hyperparameters that characterise the \emph{a priori} values the flux can take. The prior distribution converges towards an uninformative uniform prior as the prior variances increase.

We assume a Gaussian prior also for the metabolite changes 
$$\dx | \m_{\dx}, \bs_{\dot{x}} \sim \N(\mathbf{m}_{\dot{x}}, \Sigma_{\dot{x}}),$$
where $\mathbf{m}_{\dot{x}} \in \R^M$ are the \emph{a priori} mean accumulations or depletions of metabolite species. The diagonal covariances $\Sigma_{\dot{x}} = \diag(\bs_{\dot{x}}^2) = \diag( \sigma_{\dot{x}_1}^2, \ldots, \sigma_{\dot{x}_M}^2)$ encode the variances around prior metabolite changes. In strict steady-state, the prior for metabolite change becomes Dirac's delta function at zero. By increasing the variances $\bs_{\dot{x}}^2$ we can relax the steady-state assumption on individual metabolites, and encode allowance for accumulations or depletions of them.

The joint distribution of fluxes $\v$ and metabolite changes $\dx$ can now be stated as a joint multivariate Gaussian distribution
$$\begin{bmatrix} \v \\ \dx \end{bmatrix} = \begin{bmatrix} \v \\ S\v \end{bmatrix} \sim \N\left( \begin{bmatrix} \mathbf{m}_v \\ S \mathbf{m}_v \end{bmatrix}, \begin{bmatrix} \Sigma_v & \Sigma_v S^T \\ S \Sigma_v & S \Sigma_v S^T \end{bmatrix} \right),$$
that encodes the exact\footnote{We add small numerical tolerance $\kappa I$ within the inverse to ensure invertibility of the matrix.} relation $S\v = \dx$. The conditional distribution of fluxes given a specific realisation of metabolite changes $\dx$ (e.g. $\0$) is then from standard Gaussian identities
\begin{align*}
\v | \dx &\sim \N\Big( \mathbf{m}_v + \Sigma_v S^T (S \Sigma_v S^T)^{-1} (\dx - S\mathbf{m}_v), \\
& \qquad\qquad \Sigma_v -  \Sigma_v S^T (S \Sigma_v S^T)^{-1} S \Sigma_v \Big) \\
 &\sim \N ( \mathbf{m}_v + A (\dx - S\mathbf{m}_v), \Sigma_v - A S \Sigma_v )
\end{align*}
where $A = \Sigma_v S^T (S \Sigma_v S^T)^{-1}$. Since we do not in general have access to exact metabolite change values $\dx$, we marginalise the conditional flux distribution over the change prior distribution $p(\dx)$ resulting in 
\begin{align}\label{eq:prior}
p(\v | \bs_v, \bs_{\dot{x}}) &= \int p(\v | \dx) p(\dx) d \dx \nonumber \\
  &= \N(\v | \bmu, C),
\end{align}
where  $\bmu = \m_v + A (\m_{\dot{x}} - S \m_v)$ and $C = \Sigma_v - A S \Sigma_v + A \Sigma_{\dot{x}} A^T$.

\subsection{Conditioning the model with observations}
\label{sec:condition}

Assume we have access to noisy observations $\y_o = \v_o + \varepsilon$ from a subset of \emph{observed} fluxes $\v_o \subset \v$. The observations can be empirical measurements, 13C flux estimations, or flux hypotheses determined by the user. We assume independent additive Gaussian noise $\varepsilon_i \sim \N(0, \omega_i^2)$ with variances collected in a matrix $\Omega_o = \diag(\omega_1^2, \ldots, \omega_{N_{obs}}^2)$, and hence the likelihood of observed fluxes is
$$p(\y_o | \v_o, \bo_o) = \N(\y_o | \v_o, \Omega_o).$$

The joint distribution of \emph{all} fluxes $\v$ and noisy flux observations $\y_o$ is now
$$\begin{bmatrix} \v \\ \v_o + \varepsilon \end{bmatrix} = \begin{bmatrix} \v \\ \y_o \end{bmatrix} \sim \N\left( \begin{bmatrix} \bmu \\ \bmu_o \end{bmatrix}, \begin{bmatrix} C_{NN} & C_{No} \\ C_{oN} & C_{oo} + \Omega_o \end{bmatrix} \right),$$
which gives a conditional distribution of all fluxes given the observations as
\begin{equation}\label{eq:conddist}
\v | \y_o \sim \N( \bmu + C_{No} C_y^{-1} (\y_o - \bmu_o), C - C_{No} C_y^{-1} C_{oN}),
\end{equation}
where $C_y = C_{oo} + \Omega_o$ is the noisy covariance, $C_{NN}$ is the full $(N \times N)$ covariance matrix, $C_{No} = C_{oN}^T$ is the $(N_{obs} \times N)$ covariance matrix between observed fluxes and all fluxes, and $C_{oo}$ is the $(N_{obs} \times N_{obs})$ covariance matrix between observed fluxes. Note that observed fluxes are in both $\v$ and in $\v_o$. Note also that the model works with no observations at all as the conditional distribution in Eq.~(\ref{eq:conddist}) reduces back to the prior in Eq.~(\ref{eq:prior}). 

Finally, we add the flux upper and lower bounds by truncating the distribution with the known flux lower $lb$ and upper $ub$ bounds resulting in the final truncated normal \emph{flux posterior}
$$\v | \y_o \sim \TN(\bmu + C_{No} C_y^{-1} (\y_o - \bmu_o), C - C_{No} C_y^{-1} C_{oN}, lb, ub),$$
The posterior encodes the distribution of bounded fluxes that are compatible with the flux observations, flux priors, and where steady-state applies according to the tolerances determined by the steady-state prior means $\mathbf{m}_{\dot{x}}$ and variances $\bs_{\dot{x}}^2$.

The derived flux posterior is an unimodal truncated multivariate normal (TMVN) distribution where flux dependencies are represented through the covariance matrix $C$, which encodes all flux relationships with high rank. The flux posterior as a whole characterises the distribution of all valid flux vectors. The main characterisations of interest are the individual flux distributions (the marginals) and flux combination distributions (multi-variate marginals). Marginals of TMVN's are not analytically tractable, nor are they TMVN distributions \citep{horrace2005}. We resort to MCMC sampling from the TMVN flux distribution to reveal individual flux, flux pair or flux group distributions.

\subsection{Gibbs sampling truncated MVN's}

A recent review summarises sampling approaches for Truncated MVN's \citep{altmann2014}. The conditionals of TMVN's are still TMVNs \citep{horrace2005}, which has lead to many Gibbs based samplers \citep{geweke1991,kotecha1999,horrace2005,emery2014,li2015}, while also HMC samplers \citep{pakman2014} have been proposed, while elliptical slice samplers would also fit well to the problem \citep{murray2010}. We experimented with the three main approaches, and found out that Gibbs sampling has consistently the best performance in genome-scale metabolic models up to 4000 fluxes (data not shown). In the remainder of the paper we apply Gibbs sampling.

To sample the distribution $\v \sim \TN(\bmu,C,lb,ub)$ we begin by transforming it into whitened domain by Cholesky decomposition $C = LL^T$ with transformed fluxes $\tilde \v  = L^{-1}(\v - \bmu)$ with white distribution $\tilde \v \sim \TN(\0,I,\mathbf{a},\mathbf{b})$, where $\mathbf{a} = lb - L\bmu$ and $\mathbf{b} = ub - L \bmu$. We sample from the univariate conditional distributions 
\begin{align}
\label{eq:gibbs}
\tilde{v}_i | \tilde \v_{-i} \sim \TN(0,1,a(\tilde \v_{-i}), b(\tilde \v_{-i})),
\end{align}
which is a standard Normal with bounds in the white domain are (We refer to \citet{li2015} for detailed explanation):
\begin{align*}
a(\tilde \v_{-i}) &= \max\left\{ \max_{j: L_{ji} > 0} \frac{a_j - L_{j,-i} \tilde{\v}_{-i} }{L_{ji}} , \max_{j: L_{ji} < 0} \frac{b_j - \mathbf{l}_{j,-i} \tilde{\v}_{-i} }{L_{ji}}  \right\} \\
b(\tilde \v_{-i}) &=  \max\left\{ \min_{j: L_{ji} > 0} \frac{b_j - \mathbf{l}_{j,-i} \tilde{\v}_{-i} }{L_{ji}} , \min_{j: L_{ji} < 0} \frac{a_j - \mathbf{l}_{j,-i} \tilde{\v}_{-i} }{L_{ji}}  \right\}.
\end{align*}
The bounds are functions of the remaining (whitened) fluxes $\tilde \v_{-i}$ and need to be updated after each change to flux values. We iteratively update each whitened flux $\tilde{v}_i$ by sampling a new value from the conditional distribution $\tilde{v}_i | \tilde \v_{-i}$ with the minimax tilting method \citep{botev2016}, which we found out to outperform the alternative Chopin's algorithm \citep{chopin2011}. Finally, we transform the whitened variables back into original domain by $\v = L \tilde{\v} + \bmu$.

\begin{table}[t]
\begin{tabular}{lrrrrr}
     &  & & & \multicolumn{2}{c}{Runtime} \\
    Organism & Model & $N$ & $M$ & thin $100$ & thin $1000$ \\
    \hline
    \emph{E. coli}        & core    & $95$   & $72$   & $2$ min & $20$ min \\
    \emph{E. coli}        & iJR904  & $1075$ & $761$   & $2$ hr & $1$ d $9$ hr  \\
    \emph{E. coli}        & iAF1260 & $2382$ & $1668$  & $7$ hr & $4$ d $12$ hr   \\
    \emph{E. coli}        & iJO1366 & $2583$ & $1805$  & $9$ hr & $4$ d $16$ hr  \\
    \emph{B. subtilis}    & iYO844  & $1250$ & $992$   & $3$ hr & $1$ d $11$ hr  \\
    \emph{C. acetobutylicum} & Wallenius (2013) & $592$  & $444$   & $23$ min & $3$ hr \\
    \emph{S. cerevisiae}  & iMM904  & $1577$ & $1226$  & $3$ min & $2$ d  \\
    \emph{S. cerevisiae}  & 7.6     & $3493$ & $2220$  & $10$ hr & $5$ d $15$ hr \\
    \emph{T. reesei}      & CORECO  & $4008$ & $3292$  & $8$ hr & $4$ d $15$ hr \\
    \hline \\ 
\end{tabular}
\caption{Metabolic models analysed by Bayesian flux analysis by sampling $500$ samples from the flux posteriors. The runtime is shown for thinning rate $100$ and $1000$.}
\label{tab:models}
\end{table}

We notice that the method of \citet{li2015} can be further optimised by running multiple chains in parallel by considering a flux sample matrix $\tilde{V} = (\tilde{\v}^1, \ldots, \tilde{\v}^{N_c})$ containing whitened flux vector chains $\tilde{\v}^c$. The bound function is then represented as
\begin{align*}
\mathbf{a}(\tilde V_{-i}) &= \max\left\{ \max_{j: L_{ji} > 0} \frac{\mathbf{a} - L_{-i} \tilde{V}_{-i}}{L_{ji}} , \max_{j: L_{ji} < 0} \frac{\mathbf{b} - L_{-i} \tilde{V}_{-i} }{L_{ji}}  \right\} \\
\mathbf{b}(\tilde V_{-i}) &=  \max\left\{ \min_{j: L_{ji} > 0} \frac{ \mathbf{b} - L_{-i} \tilde{V}_{-i} }{L_{ji}} , \min_{j: L_{ji} < 0} \frac{ \mathbf{a} - L_{-i} \tilde{V}_{-i} }{L_{ji}}  \right\},
\end{align*}
where $\tilde{V}_{-i}$ is the sample matrix $\tilde{V}$ without the $i$'th row, $L_{:,-i}$ is the Cholesky matrix without the $i$'th column. By sampling several chains in parallel with one CPU node we can utilise the full bandwidth of the CPU. This is especially useful for Gibbs sampling.

\subsection{Sampling the flux posterior}

\begin{figure}[t!]
\centering \includegraphics[width=\columnwidth]{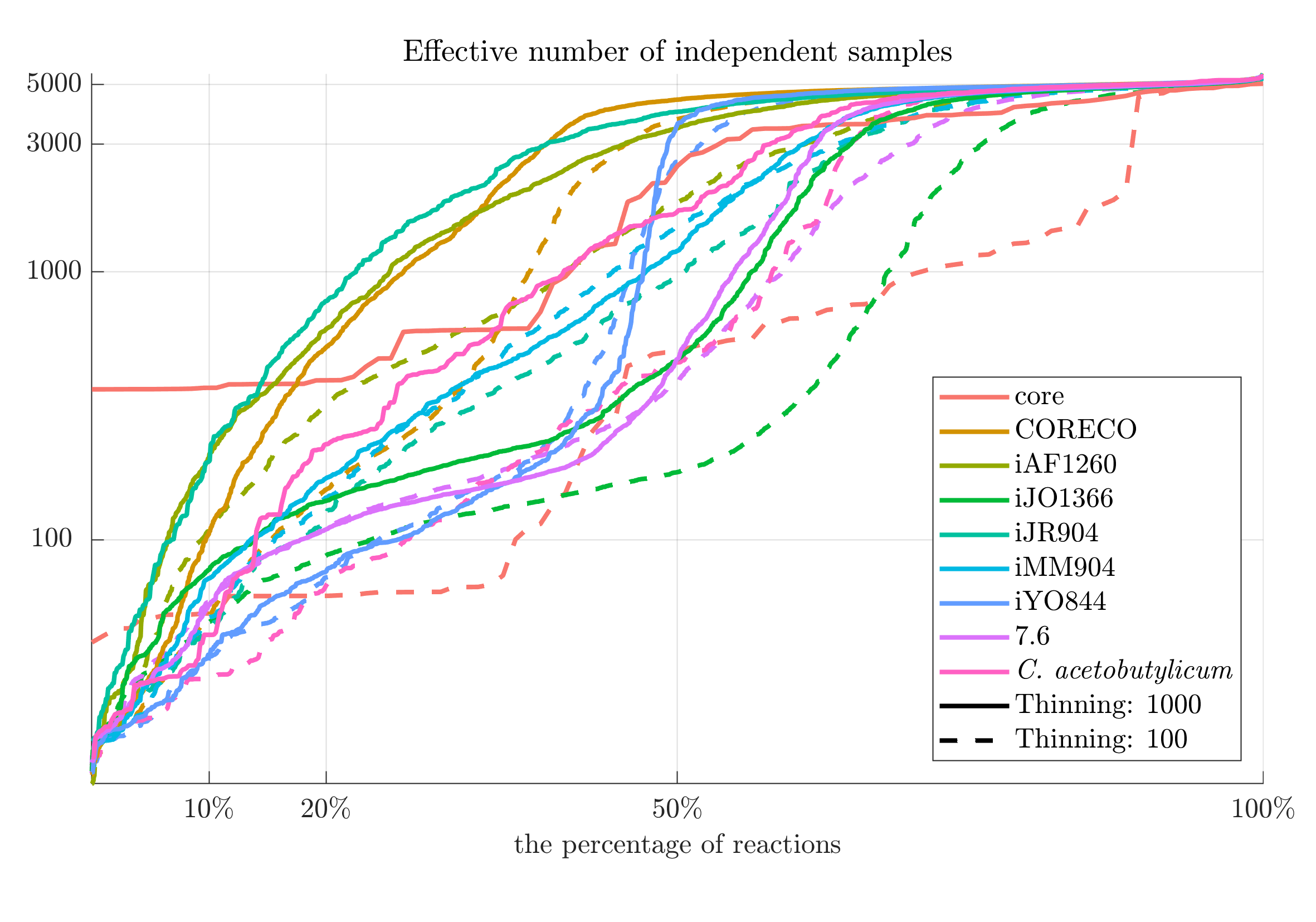}
\caption{The effective number of independent simulation draws for a individual fluxes for a subset of models from Bayesian flux analysis with different thinning parameters. The x-axis corresponds to individual fluxes sorted based on the effective number of samples.}
\label{fig:neff}
\end{figure}

We set the initial flux vector $\v^{(0)} = \v_{\text{MAP}} =  \argmax_{\v} p(\v)$ to the \emph{maximum a posteriori} (MAP) of the truncated normal distribution, which we compute using quadratic programming. The truncated normal distribution is unimodal, and hence we begin sampling from the mode of the distribution providing efficient optimization. 

The fluxes can be arranged into distinct bounded fluxes $\v_b$ and unbounded fluxes $\v_u$, where $\v = (\v_b, \v_u)^T$. The conditional distribution of a truncated normal is still a truncated normal \citep{horrace2005}. We only need to sample the bounded fluxes $\v_b$ with Gibbs MCMC, and afterwards the distribution of unbounded fluxes $\v_u$ conditioned on the bounded flux samples $\v_b$ can be drawn from untruncated normal as $$\v_u | \v_b \sim \N( \bmu_u + C_{ub} C_{bb}^{-1} (\v_b - \bmu_b), C_{uu} - C_{ub} C_{bb}^{-1} C_{bu} ),$$ where we arrange $\bmu = (\bmu_b, \bmu_u)^T$ and $C = \begin{pmatrix} C_{bb} & C_{bu} \\ C_{ub} & C_{uu} \end{pmatrix}$.

We implement the MCMC sampling in Matlab. We run multiple independent Markov chains in a vectorised form. By default we run $N_c = 10$ chains of $N_s = 500$ flux samples for a total of $5000$ flux vector samples. We use thinning and only accept every $N_t = 100$'th flux sample. The MCMC chains have converged if successive samples are uncorrelated, chains are indistinguishable and have effectively forgotten the initial value. Convergent chains indicate that the MCMC sampler has characterised the whole flux posterior. We use \emph{potential scale reduction factor} $\hat{R}$ to approximate convergence \citep{gelman1992}. An optimal value of $\hat{R}=1$ indicates convergence, while values $\hat{R} < 1.1$ are considered sufficient for convergence. We also compute the effective number of samples $N_{\mathrm{eff}}$ per flux \citep{gelman2013}. 

We assume that flux means $\mathbf{m}_v$ are fixed to either zero or the lower bound of each respective flux. The model has then two main hyperparameters $\bs_v$ and $\bs_{\dot{x}}$ that affect the posterior. The $\bs_{\dot{x}}$ determines how much the mass balance can be relaxed, and can be set according to the prior knowledge of the modeller. To enforce mass balance, a small value such as $\bs_{\dot{x}} = 0.001$ should be chosen. The prior flux variance $\bs_v^2$ determines how much fluxes are driven towards zero \emph{a priori}, but also should be set to sufficiently high value not to exclude possibly high fluxes. In practise we set the variance $\sigma_{v_i}^2 = 100^2$ for all fluxes $v_i$.

\subsection{Sampling FBA solutions}

The presented Bayesian model is a metabolic flux analysis (MFA) model designed to characterise the global flux configurations $\v \sim p(\v | \y_{obs})$ compatible with mass balance assumption, observations, and bounds. The method can as well be applied as an flux balance analysis (FBA) method, where a target function -- such as biomass reaction -- is maximised. For FBA mode we first find the standard FBA solution target flux $\v_{\text{target}}^{\text{FBA}}$ with linear programming, and encode it as a flux observation $\y_{\text{target}} \pm \omega_c^2$, where the variance determines how closely we sample from the maximal target. By default, we set the standard deviation to $0.1\%$ of the target flux. 
To run maximum growth Bayesian FBA we would set an observation for the biomass pseudoreaction to the classical FBA maximum growth value and condition the model with the pseudo-measurement as in Sec.~\ref{sec:condition}.

\begin{figure}[t]
\centering \includegraphics[width=0.99\columnwidth]{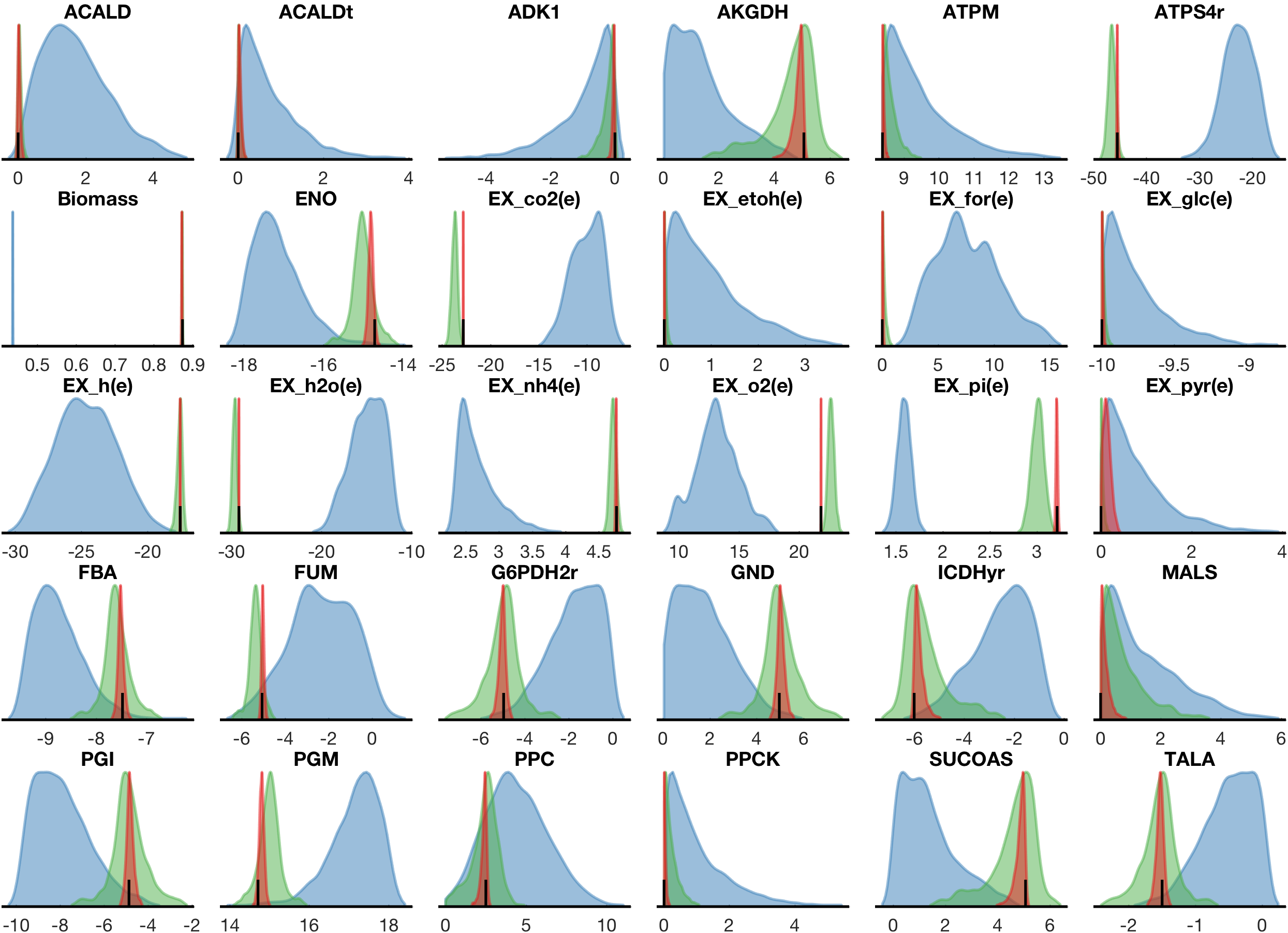}
\caption{Posterior flux distributions of {\em E.\ coli} core model. The blue color indicates fluxes in $50\%$ growth, the green color maximal growth, the red color maximal growth with $9$ exchange fluxes specified, and the conventional FBA solution is a black line.}
\label{fig:flux}
\end{figure}

\section{Results}

We first perform \emph{in silico} experiments to highlight the capabilities of the Bayesian FBA and MFA models in Sections~\ref{sec:sampling}--\ref{sec:couplings}. Our goal is to compare the computational approach against the conventional FBA and FVA methods, and to showcase the method's \emph{in silico} performance in various metabolic models. The main experiment of this paper is application of the Bayesian flux analysis to the \textsuperscript{13}C analysis of the C. acetobutylicum in Section~\ref{sec:aceto}, where we can elucidate fluxes on a genome-scale from a small set of intracellular flux measurements.

\subsection{In silico metabolic models}
\label{sec:sampling}

Table \ref{tab:models} indicates the stoichiometric models that were considered. We considered four organisms, seven genome-scale metabolic models and one core model. All models were downloaded from the BiGG database\footnote{http://bigg.ucsd.edu}. For all models we run the Bayesian model in FBA mode -- by specifying a growth target -- with standard exchange flux measurements included as bounds. We sampled $10$ chains of $500$ flux vector samples from the full space containing all intracellular and extracellular fluxes. These $5000$ flux vectors represent all possible flux configurations compatible with the experimental setting. The sampling thinning parameter determines how uncorrelated successive MCMC samples are. We applied thinning values of $100$ and $1000$, with linear effect on the running time. 

The effective number of independent simulation draws for all models from Bayesian flux analysis with different thinning parameter are shown in Figure \ref{fig:neff} using the potential scale reduction factor. The x-axis corresponds to individual fluxes sorted based on the effective number of samples. The number of reactions is different for different models. In all cases majority of the fluxes have over $100$ effective number of independent samples, which indicates that the samples represent the flux posterior well. A minority of the fluxes have low effective sample sizes. These are usually central branching fluxes that are highly dependent across the genome-scale metabolism, and hence converge slowly. The thinning parameter has a large effect on some models (core, iJR904, iAF1260, iJO1366) whereas for some other models there are not much change (CORECO, iYO844, iMM904, 7.6).

\subsection{Bayesian FBA and MFA}
\label{sec:marginals}

We illustrate the characteristics of the Bayesian model using \emph{E. coli} central carbon metabolism model\footnote{BiGG model \texttt{e\_coli\_core}}. The model contains $95$ fluxes and $72$ metabolites. It should be noted that the model was not further constrained and do not represent the native \emph{ E. coli} strain as such, while it allows e.g. carbon fixation. The model was rather used to theoretically compare our modelling approach to conventional FBA methodology. The conventional FBA solution achieves a growth flux $v_{\text{growth}}^{\text{FBA}} \approx 0.873$ by limiting the glucose exchange flux with a $lb_{\text{glc}} = -10$. We consider three cases of Bayesian analysis: (i) $50\%$ growth by defining only the biomass flux observation $0.436 \pm 0.0043$, (ii) maximal growth scenario by defining the biomass flux observation $0.873 \pm 0.0087$, and (iii) maximal growth with additional observations for $9$ key exchange fluxes: \texttt{GLC}, \texttt{O2}, \texttt{CO2}, \texttt{H2O}, \texttt{H+}, \texttt{HPO4}, \texttt{SO4}, \texttt{NH4} and \texttt{ethanol} that were all set to their conventional FBA solutions with a standard deviations of $0.01$. In all three experiments the remaining fluxes were free with only a prior distribution with a standard deviation of $100$ specified. We defined a nearly strict steady-state by defining $\sigma_{\dot{x}} = 0.01$. We sample a total of $5000$ flux vectors with the Gibbs sampler using $10$ chains and $500$ samples each. We use 1-dimensional kernel density estimates as proxies of marginal flux posterior distributions. The small jaggedness of the distributions are an artefact from the MCMC sampling. By considering longer chains these would eventually smoothen out. 

Figure \ref{fig:flux} shows the flux distributions of $30$ fluxes. The blue color indicates the $50\%$ growth flux distributions, the green color the maximal growth distributions, the red color maximal growth with exchange fluxes specified, and the conventional FBA is shown with a black line. The Bayesian distributions represent the space of all allowed steady-state flux configurations given the observations and target function. Figure \ref{fig:flux} shows that maximal growth can still support a large variance in many fluxes, with the FBA point estimate misleading by only considering one flux configuration. Similarly to conventional FVA our approach elucidates directly the possible variance in a given flux. For instance, the pentose-phosphate pathway flux G6PDH2r: D-Glucose 6-phosphate + NADP $\Leftrightarrow$ 6-Phosphogluconolactone + H$^{+}$ + NADPH can vary between $-8$ and $-2$ in maximal growth. The conventional FBA yields zero flux for glyoxylate cycle flux MALS: Acetyl-CoA + Glyoxylate + H$_2$O $\Rightarrow$ CoA + H$^{+}$ + Malate, while the flux space indicates that values up to $4$ are possible.

The red distributions indicate how the intracellular fluxes get more and more specified as the model is better specified by inclusion of exchange measurements. Variance of almost all fluxes reduces by more than half. For instance, the flux FUM: Fumarate + H$_2$O $\Leftrightarrow$ Malate is specified to a range of $[-5.1, -4.9]$ from a range of $[-6.5, 4.4]$ without exchange measurements.

The blue color indicates the cellular flux state when the cell is only growing at $50\%$ of the maximum growth rate. Most fluxes have a higher variance in this scenario. Interestingly the glucose intake is still kept at a relatively high rate. Instead of biomass production, the excess carbon from glucose can be diverted to other carbon sinks, such as formate and ethanol production. The ethanol and formate effluxes can grow up to 3 and 15, respectively. The carbon dioxide exchange decreases by over half into a range of $[-15, -3]$ from maximal growth exchange range of $[-25, -23]$. 

\begin{figure}[t!]
\centering \includegraphics[width=0.99\columnwidth]{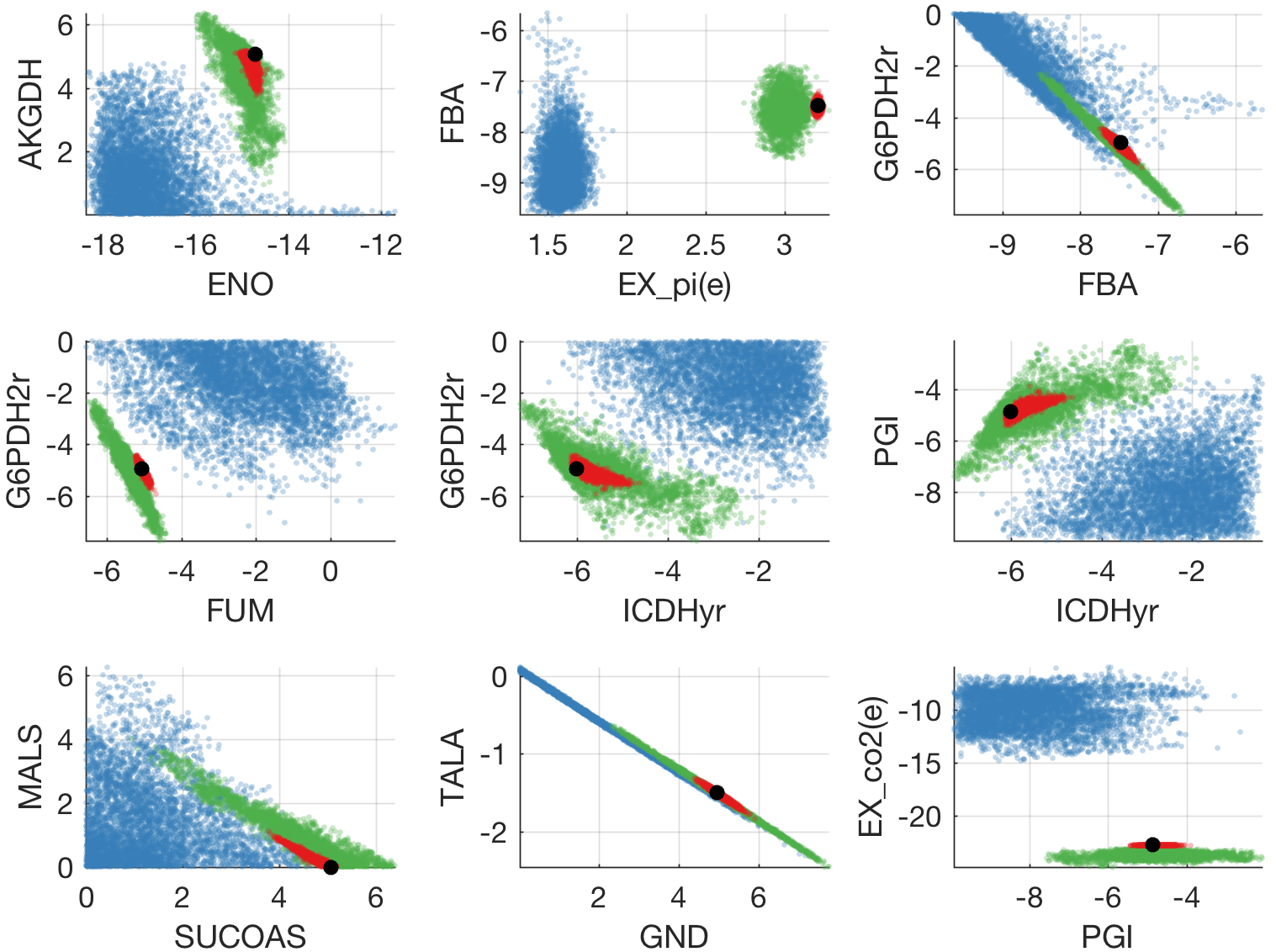}
\caption{Examples of flux covariance distributions of \emph{E.\ coli} core network. Blue points represent $50\%$ growth, green points maximal growth, red points maximal growth with $9$ exchange fluxes specified, and the conventional FBA solution is a black dot. Scatter plots represent pair-wise (2-dimensional) marginal posterior distributions as obtained from the MCMC samples.}
\label{fig:cov}
\end{figure}

\subsection{Flux couplings}
\label{sec:couplings}

The flux variations are in general not independent from each other. To understand the intracellular flux space, we need to consider higher-order flux dependencies. The flux sample covariances indicate \emph{flux couplings}, where the variation in one flux is constrained by other fluxes. Figure \ref{fig:cov} highlights $9$ example flux pair patterns out of the total of $\frac{95 \cdot 94}{2} = 4465$ in the core model.  Blue points represent $50\%$ growth, green points maximal growth, red points maximal growth with $9$ exchange fluxes specified, and the conventional FBA solution is a black dot.

\begin{figure*}[t]
\centering \includegraphics[width=1.05\textwidth]{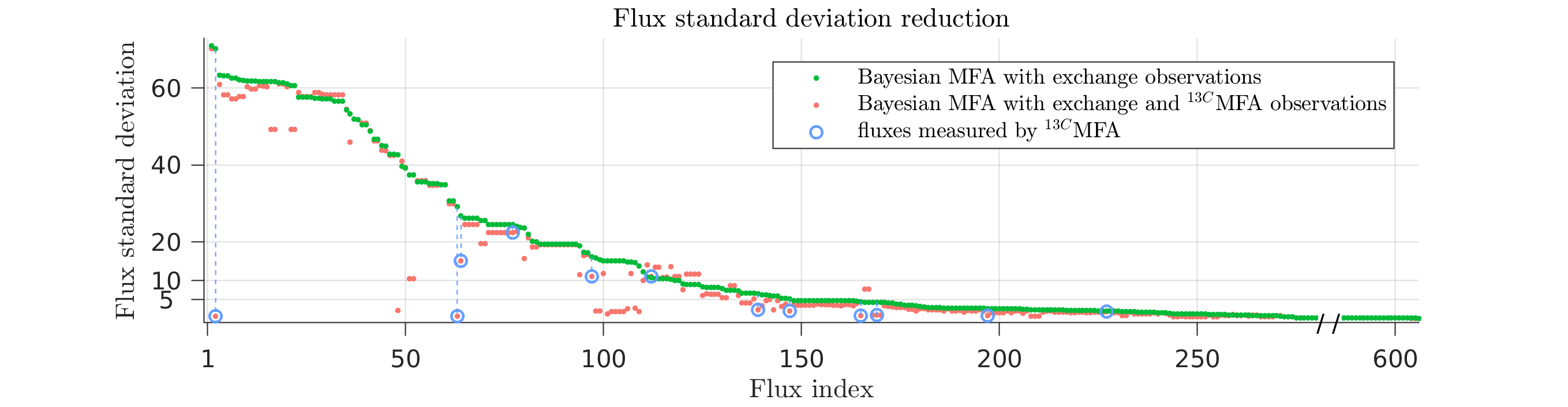}
\caption{Global flux standard deviation reduction due to addition of twelve \textsuperscript{13C}MFA internal flux measurements in glucose limited condition. The green points indicates the standard deviation of fluxes given only exchange measurements. The red points indicate the corresponding standard deviations after inclusion of twelve \textsuperscript{13C}MFA intracellular flux measurements. The blue circles highlight the \textsuperscript{13C}MFA measured fluxes. }
\label{fig:fluxred}
\end{figure*}

The flux covariations become consistently more constrained while traversing from the loose $50\%$ growth model (blue) towards $100\%$ maximal growth (green). By measuring the exchange fluxes (red), we can already pinpoint most flux patterns very closely around the theoretically optimal flux pattern as defined by the conventional FBA (black).

\begin{figure*}[t!]
\centering \includegraphics[width=\textwidth]{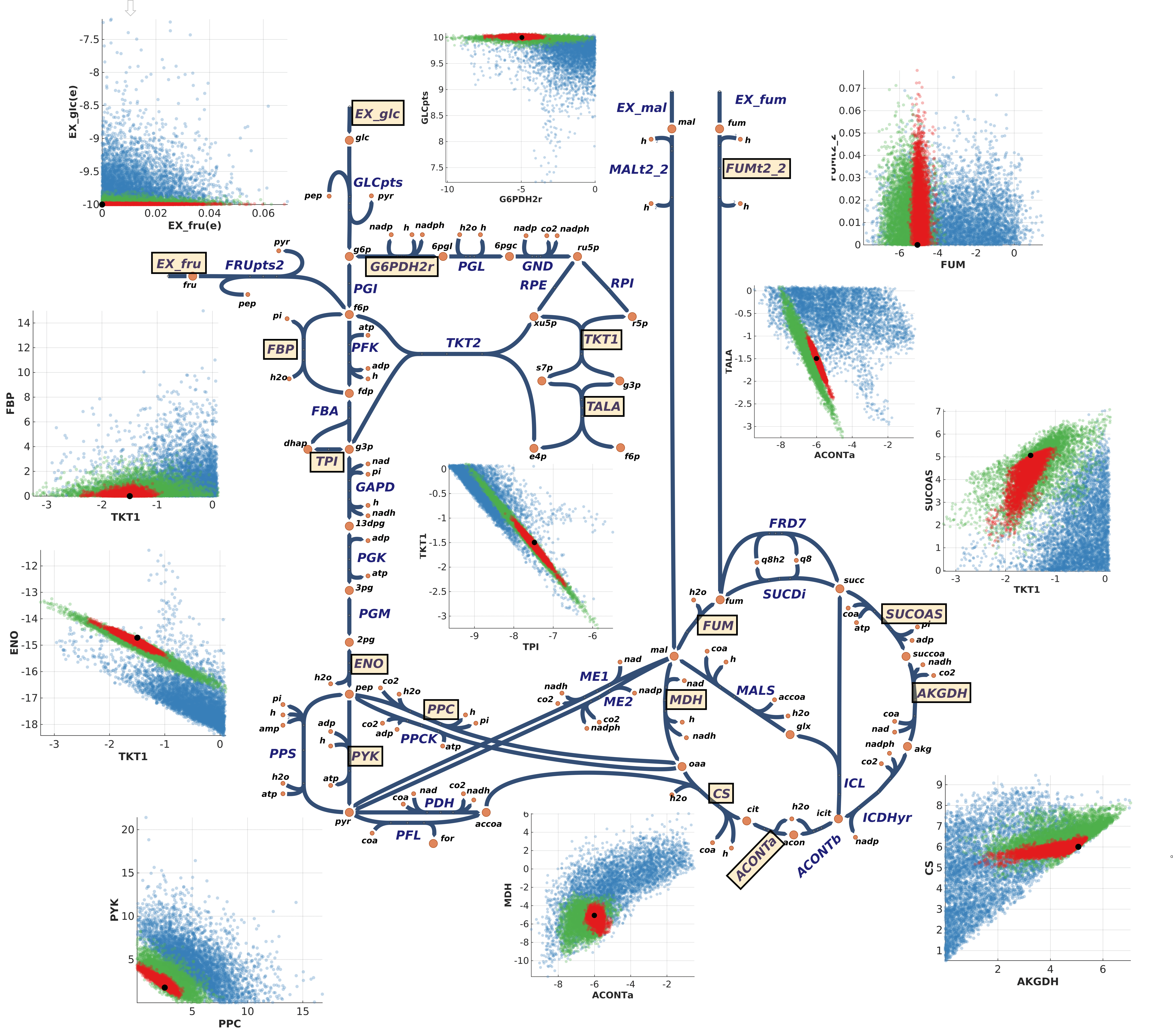}
\caption{Pair-wise marginal posterior fluxes presented together the the metabolic network map. The visualized fluxes are highlighted. Blue points represent flux values in $50\%$ growth, green points in maximal growth, red points in maximal growth with $9$ exchange fluxes specified, and the conventional FBA solution is a black dot.}
\label{fig:map_pairs}
\end{figure*}

Multiple patterns of covariation can be immediately identified. There is an exact coupling between pentose-phosphate pathway reactions GND: 6-Phospho-D-gluconate + NADP $\Rightarrow$ CO2 + NADPH + D-Ribulose 5-phosphate and TALA: Glyceraldehyde 3-phosphate + Sedoheptulose 7-phosphate $\Leftrightarrow$ D-Erythrose 4-phosphate + D-Fructose 6-phosphate, as expected from the stoichiometry. Glycolysis related FBA: D-Fructose 1,6-bisphosphate $\Leftrightarrow$ Dihydroxyacetone phosphate + Glyceraldehyde 3-phosphate and pentose-phosphate pathway related G6PDH2r have also a strong, but not exact, negative correlation. The flux PGI: D-Glucose 6-phosphate $\Leftrightarrow$ D-Fructose 6-phosphate and carbon dioxide exchange have no correlation, but the maximal growth still pinpoints to a carbon dioxide exchange value of approximately $-21$. The dependency of glyoxylate cycle related MALS and TCA cycle related SUCOAS: ATP + CoA + Succinate $\Leftrightarrow$ HPO$_4^{-2}$ + ADP + Succinyl-CoA on maximal growth requirement can be observed in Figure \ref{fig:cov}. Under maximal growth a negative correlation between the two fluxes emerges. The conventional FBA solution pinpoints an optimal values as zero MALS with SUCOAS around $5$, while the Bayesian model reveals that MALS can still have a flux value of around $4$ as long as SUCOAS tends towards $1.5$ simultaneously.

The patterns of G6PDH2r and FBA and FUM indicate a linear inequality for these fluxes. Especially with FUM this is natural since the pentose-phosphate pathway flux G6PDH2r limits the TCA cycle flux FUM. The same effect is also seen with G6PDH2r and ICDHyr: Isocitrate + NADP $\Leftrightarrow$ 2-Oxoglutarate + CO$_2$ + NADPH, another TCA cycle flux.


To get more insight into the biology behind the flux couplings, the flux pair patterns can also be illustrated in the metabolic network (Figure \ref{fig:map_pairs}). Figure \ref{fig:map_pairs} shows the samples of the flux distributions for several example pairs of reactions. These scatter plots indicate the dependency of the flux configurations between two reactions across the reaction.  
There is natural correlation between adjacent or subsequent fluxes but also correlation between fluxes in different pathways, such as glycolysis and TCA cycle (See the \texttt{SUCOAS} and \texttt{TKT1} pair).




\subsection{Intracellular flux elucidation of \emph{C. acetobutylicum}}
\label{sec:aceto}


We consider the results obtained from \textsuperscript{13C}MFA of \emph{Clostridium acetobutylicum} grown in chemostat, i.e.\ in continuous cultivation maintaining steady-state, with reference condition, glucose limited condition and butanol stimulus with the goal of inferring the internal fluxes. We effectively repeat the study of 
\citet{wallenius2016}, where FBA and FVA were performed and constrained on 12 intracellular fluxes determined by \textsuperscript{13C}MFA and 7 exchange fluxes. The model for \emph{Clostridium acetobutylicum} consists of 451 metabolites and 604 reactions, and is given as an .xml file in the supplement of \citet{wallenius2016}. 


The data from different measurements in glucose limited condition are shown in Table \ref{tab:fluxes}. For the reference and the butanol stimulated conditions, see Supplementary Tables 2 and 3. The internal fluxes are obtained from \textsuperscript{13C}MFA analysis, whereas the exchange fluxes were measured by chromatographical methods or transferred from the \textsuperscript{13C}MFA results. Flux values were normalized to the specific growth rate which was given the value of $1$, except for the reference condition, the measured growth is $0.95$. Exchange fluxes measured from the cultivations were given to Bayesian MFA as mean $\v_o$ and standard deviations $\Omega_0=0.05 \cdot \v_o$ and fluxes obtained from \textsuperscript{13C}MFA were given as ranges. In all Bayesian MFA experiments, the steady state relaxation was $\sigma_{\dot{x}} = 0.01$. Finally, 500 samples were drawn from the posterior with thinning 1000.

To study the FBA's, FVA's and BMFA's performance to predict the measured distributions of metabolic fluxes, three sets of models were generated: (A) a model with reaction {\em directions} for the 12 \textsuperscript{13C}MFA determined internal fluxes set according to the data (bounds in table \ref{tab:fluxes}
), (B) a set of 12 models where each reaction among 12 \textsuperscript{13C}MFA determined internal fluxes is unconstrained at a time (the reaction direction is still constrained) while the rest 11 fluxes are constrained according to the measurements (leave-one-out), and (C) a model with all 12 \textsuperscript{13C}MFA determined internal fluxes constrained to their measured values. In all three cases we constrain the model with measured values for the 6 exchange reactions and the measured growth. 
For the (B) set of models, we test how well we can predict the \textsuperscript{13C}MFA determined flux interval for the leave-one-out reaction by comparing the prediction with \textsuperscript{13C}MFA determined flux. For each set of models, the standard MFA with Taxicab penalty, FVA, and Bayesian MFA were performed. The MFA and FVA were performed by The Cobra Toolbox's \texttt{optimizeCbModel} and \texttt{fluxVariability} functions by maximizing growth with the growth lower and upper bounds set to the measured value $1$. 

\begin{table}[b]
    \resizebox{\columnwidth}{!}{
    \begin{tabular}{lcc r@{\hspace{1pt}}c@{\hspace{1pt}}l c c | c c | c c }
    \hline
    Reaction & KEGG ID & Bounds & \multicolumn{3}{c}{Glucose limited} & \multicolumn{6}{c}{} \\
    \hline
    \multicolumn{12}{l}{Exchange fluxes} \\
    \hline
    $^\dagger$Glucose exchange   & C00031 & $\le 0 \hspace{9pt}$ & $-73.3 $ & $\pm$ & $ 3.7$  & \multicolumn{6}{c}{}\\
    $^\ddagger$Acetate exchange  & C00033 & $\hspace{9pt} 0 \le$ & $[12.96 $ & $\dd$ & $13.016]$ & \multicolumn{6}{c}{}\\
    $^\dagger$Acetone exchange   & C00207 & $\hspace{9pt} 0 \le$ & $12.5 $ & $\pm$ & $ 0.06$  &\multicolumn{6}{c}{}\\
    $^\ddagger$Butanol exchange  & C06142 & $\hspace{9pt} 0 \le$ & $[29.62 $ & $\dd$ & $30.67]$ &\multicolumn{6}{c}{}\\
    \,\,Butyrate exchange        & C00246 & $\hspace{9pt} 0 \le$ & $[0 $ & $\dd$ & $3.23]^\ddagger$ &\multicolumn{6}{c}{}\\
    $^\dagger$Ethanol exchange   & C00469 & $\hspace{9pt} 0 \le$ & $6.13 $ & $\pm$ & $ 0.31$ &\multicolumn{6}{c}{}\\
    $^\ddagger$EPS production    &        & $\hspace{9pt} 0 \le$ & $[10.01 $ & $\dd$ & $10.26]$ &\multicolumn{6}{c}{}\\
    $^\dagger$Growth             &        & $\hspace{9pt}0 \le$  & $1.00 $ & $\pm$ & $ 0.05$  &\multicolumn{6}{c}{}\\
    \hline
    \multicolumn{6}{l}{} & \multicolumn{6}{c}{F1 scores \%} \\
    \multicolumn{6}{l}{} & \multicolumn{2}{c}{ex only} & \multicolumn{2}{c}{LOO} & \multicolumn{2}{c}{ex + 13C} \\
    \cmidrule(lr){7-8} \cmidrule(lr){9-10} \cmidrule(lr){11-12}
    \multicolumn{6}{l}{} & FVA & BMFA & FVA & BMFA & FVA & BMFA \\
    \hline
    Malate DHase          	&	 R00342 	&	      	&	 $[-112 $ 	&	 $\dd$ 	&	 $-5.94]$                  	&	\textbf{19}	&	7	&	\textbf{65}	&	19	&	\textbf{68}	&	23	\\
    3P-glycerate DHase    	&	 R01513 	&	 $\hspace{9pt} 0 \le$ 	&	 $[1.53 $ 	&	 $\dd$ 	&	 $4.85]$   	&	1	&	\textbf{53}	&	12	&	\textbf{59}	&	\textbf{100}	&	\textbf{100}	\\
    Acetaldehyde DHase    	&	 R00228 	&	      	&	 $[-27.4 $ 	&	 $\dd$ 	&	 $50.3]$                  	&	8	&	\textbf{56}	&	12	&	\textbf{73}	&	\textbf{95}	&	92	\\
    Triosephosphate DHase 	&	 R01061 	&	 $\hspace{9pt} 0 \le$ 	&	 $[77.2 $ 	&	 $\dd$ 	&	 $132]$    	&	10	&	\textbf{95}	&	5	&	\textbf{75}	&	63	&	\textbf{85}	\\
    Acetolactate synthase         	&	 R04672 	&	      	&	 $[-95.2 $ 	&	 $\dd$ 	&	 $99.4]$                  	&	17	&	\textbf{67}	&	17	&	\textbf{68}	&	68	&	\textbf{68}	\\
    Aspartate transaminase        	&	 R00355 	&	 $\le 0 \hspace{9pt}$ 	&	 $[-8.95 $ 	&	 $\dd$ 	&	 $-0.80]$ 	&	1	&	\textbf{32}	&	14	&	\textbf{29}	&	69	&	\textbf{69}	\\
    5,10-CH=THF hydrolyase        	&	 R01655 	&	      	&	 $[-2.24 $ 	&	 $\dd$ 	&	 $0.05]$                  	&	0	&	\textbf{3}	&	0	&	\textbf{3}	&	100	&	\textbf{100}	\\
    Malate hydrolyase             	&	 R01082 	&	 $\le 0 \hspace{9pt}$ 	&	 $[-10.2 $ 	&	 $\dd$ 	&	 $-0.78]$ 	&	2	&	\textbf{83}	&	\textbf{59}	&	54	&	62	&	\textbf{67}	\\
    Ribulose-5P epimerase         	&	 R01529 	&	      	&	 $[-4.39 $ 	&	 $\dd$ 	&	 $-1.27]$                 	&	\textbf{1}	&	0	&	\textbf{1}	&	0	&	100	&	\textbf{100}	\\
    Pyruvate carboxylase          	&	 R00344 	&	 $\hspace{9pt} 0 \le$ 	&	 $[13.6 $ 	&	 $\dd$ 	&	 $119]$    	&	19	&	\textbf{36}	&	\textbf{65}	&	33	&	\textbf{66}	&	26	\\
    Carbonate hydrolyase          	&	 R10092 	&	      	&	 $[26.2 $ 	&	 $\dd$ 	&	 $75.7]$                   	&	10	&	\textbf{57}	&	\textbf{97}	&	31	&	\textbf{100}	&	53	\\
    C-acetyl transferase          	&	 R00212 	&	 $\hspace{9pt} 0 \le$ 	&	 $[66.3 $ 	&	 $\dd$ 	&	 $154]$    	&	\textbf{16}	&	2	&	\textbf{84}	&	13	&	\textbf{100}	&	81	\\
    \hline
	Average &		&		&		&		&	 &	9	&	\textbf{41}	&	36	&	\textbf{38}	&	\textbf{83}	&	72	\\
    \end{tabular}
    }
    \caption{The 6 exchange flux, the growth and EPS production, and the 12 internal flux measurements. Measurements from the cultivations include standard deviations, while fluxes determined by \textsuperscript{13C}MFA$^*$ are given as a range. The unit used for the model fluxes is: \si{\milli\mol \per \gram}(CDW). $^\dagger$ : measured by chromatographic methods, $^\ddagger$ obtained from \textsuperscript{13C}MFA }
    \label{tab:fluxes}
\end{table}














\begin{figure*}[tbh]
     \centering
     \subfloat[][Only exchange fluxes observed]{\includegraphics[width=\textwidth]{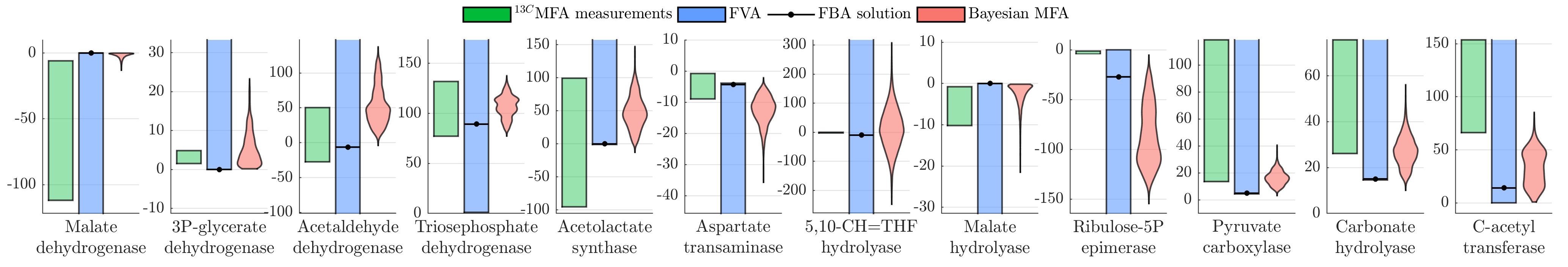}} \\
     \subfloat[][Exchange fluxes and leave-one-out internal fluxes observed]{\includegraphics[width=\textwidth]{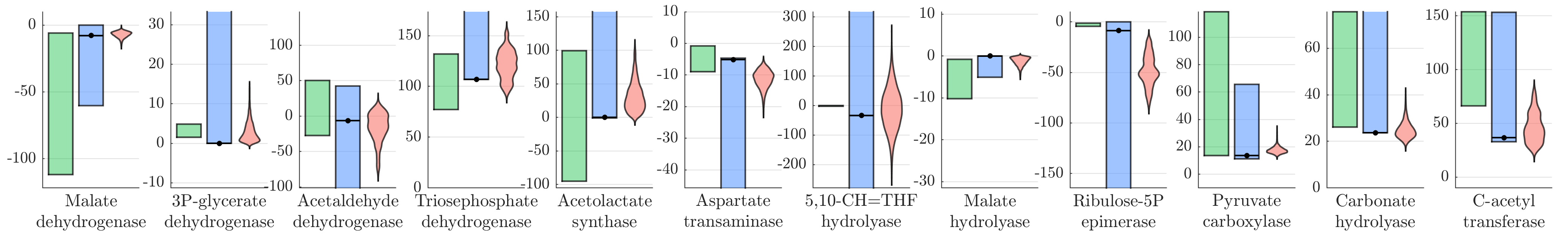}} \\
     \subfloat[][Exchange and internal fluxes obtained from \textsuperscript{13C}MFA fluxes observed]{\includegraphics[width=\textwidth]{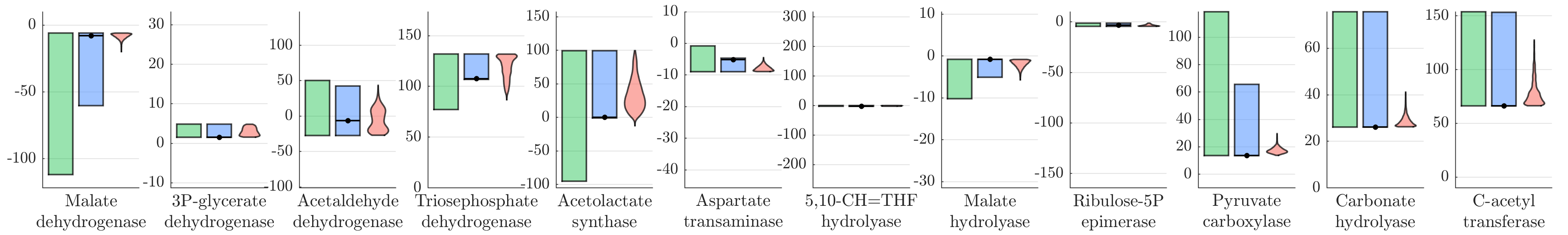}}
     \caption{For the glucose limited condition, flux distributions of the 12 internal fluxes predicted solely from exchange fluxes \textbf{(a)}, and distributions after seeing \textsuperscript{13C}MFA data in leave-one-out experiment \textbf{(b)} and after seeing all 12 internal reactions \textbf{(c)}. }
     \label{fig:clost}
\end{figure*}

We study how the flux variances from Bayesian MFA results decrease when adding  \textsuperscript{13C}MFA constraints. The Figure \ref{fig:fluxred} shows the reduction of standard deviations of flux distributions of all fluxes of \emph{C. acetobutylicum} in glucose limited condition when all 12 \textsuperscript{13C}MFA constraints are added to the model (model set C). When adding the \textsuperscript{13C}MFA constraints, the variance of most unmeasured internal fluxes decreases, demonstrating how Bayesian MFA propagates the information about 12 measured fluxes to several tens of other internal fluxes. The reduction of standard deviations of flux distributions for the reference and butanol stimulated conditions are shown in Supplementary Figures 2 and 3.



For the glucose limited experiment, the MFA, FVA and Bayesian MFA results for model sets A, B and C are shown for 12 \textsuperscript{13C}MFA determined reactions in Figure \ref{fig:clost}. To quantify the performance of FVA and Bayesian MFA to predict the \textsuperscript{13C}MFA measured range of flux values, precision, recall and the F$_1$ score were computed for each reaction and model set. The F$_1$ score values are shown in Table \ref{tab:fluxes} as percentage and the precision and recall values are shown in the Supplementary Table 1. From Table \ref{tab:fluxes} it can be concluded that the Bayesian MFA outperforms FVA for most of the reactions, especially in the leave-one-out testing. In Figure \ref{fig:clost}, for the model set A, the FVA gives a wide range of solutions whereas the distribution from Bayesian MFA is narrower and closer to the true values for almost all reactions. Figure \ref{fig:clost}b
shows the results for the model set B: for reactions 3P-glycerate dehydrogenase, Acetaldehyde dehydrogenase, Triosephosphate dehydrogenase and Acetolactate synthase the flux distribution obtained by Bayesian MFA matches the test data better. In Figure \ref{fig:clost}c
the resulting flux ranges for FVA and Bayesian MFA distributions are always within the true measured range, but the Bayesian MFA captures the probability in the flux values. Similar results are obtained for the butanol stimulated and reference conditions (See Supplementary Tables 2 and 3 Supplementary Figures 4 and 5).

\section{Discussion}

The conventional FBA formalism is a powerful framework for flux analysis that however assumes several unrealistic simplifying model approximations. Several approaches from robust flux analysis and sampling to flux variability analyses indicate the need to alleviate the approximations towards a more principled model.

We proposed the Bayesian flux analysis formalism that considers fluxes as distributions instead of point estimates. The model learns a posterior distribution of fluxes given prior information, flux measurements, upper and lower bounds and steady-state assumptions into account. The degree of belief in the measurements and steady-state can be adjusted via measurement noise variances and biological knowledge as encoded in (subjective) priors. The model characterises the complete space of possible flux configurations by modeling the uncertainties of fluxes and flux combinations. The Bayesian formalism can be seen as a drop-in replacement for deterministic flux analysis tools --- such as FBA and FVA --- at the cost of added running time necessary to properly characterise the flux distributions. The runtime can be effectively alleviated by only considering the core parts of the metabolic model or by running multiple MCMC chains in parallel.

Our results show that the conventional FBA and FVA tools provides an overly simplistic view of the flux capabilities of the cellular system under study, while the Bayesian model expresses the full variance in the flux configurations. The Bayesian model of metabolism opens doors for building flux analysis models in a Bayesian way. We will leave experimental design, knock-outs and strain design using the Bayesian modelling basis for future work. In future we expect the Bayesian formalism to provide an alternative statistical approach for majority of current FBA and MFA based tools with the benefit of rigorous uncertainty modeling and improved interpretation.

\section*{Acknowledgements}

This work has been supported by the Academy of Finland Center of Excellence in Systems Immunology and Physiology, the Academy of Finland grants no. 260403 and 299915, and the Finnish Funding Agency for Innovation Tekes (grant no 40128/14, Living Factories). \vspace*{-12pt}

\bibliographystyle{natbib}
\bibliography{refs}

\end{document}